\pdfoutput=1
\documentclass[11pt]{article}

\usepackage[]{ACL2023}

\usepackage{times}
\usepackage{latexsym}

\usepackage[T1]{fontenc}
\usepackage[utf8]{inputenc}

\usepackage{microtype}

\usepackage{inconsolata}

\usepackage{booktabs}
\usepackage{multirow}
\usepackage{multicol}
\usepackage{makecell}
\usepackage{enumitem}
\usepackage{balance}
\usepackage{graphicx}
\usepackage{amsmath}
\usepackage{amsthm}
\usepackage{amsfonts}
\usepackage{hyperref}

\newcolumntype{L}[1]{>{\raggedright\let\newline\\\arraybackslash\hspace{0pt}}m{#1}}
\newcolumntype{C}[1]{>{\centering\let\newline\\\arraybackslash\hspace{0pt}}m{#1}}
\usepackage{color}
\definecolor{red}{HTML}{FF0000}
\definecolor{blue}{HTML}{0000FF}
\definecolor{dgreen}{HTML}{228B22}
\definecolor{dblue}{HTML}{007FFF}
\DeclareRobustCommand{\bluebox}[1]{\setlength{\fboxsep}{1.0pt}\colorbox{dblue!25}{#1}}
\DeclareRobustCommand{\greenbox}[1]{\setlength{\fboxsep}{1.0pt}\colorbox{dgreen!25}{#1}}
\DeclareRobustCommand{\redbox}[1]{\setlength{\fboxsep}{1.0pt}\colorbox{red!15}{#1}}

\title{Dialogue Planning via Brownian Bridge Stochastic Process for Goal-directed Proactive Dialogue}

\author{Jian Wang\thanks{\ \ Equal contribution.}, ~
    Dongding Lin\footnotemark[1], ~
    Wenjie Li \\
  Department of Computing, The Hong Kong Polytechnic University \\
  \texttt{\{jian-dylan.wang, dongding88.lin\}@connect.polyu.hk} \\
  \texttt{cswjli@comp.polyu.edu.hk}
  }

\begin{document}
\maketitle
\begin{abstract}
Goal-directed dialogue systems aim to proactively reach a pre-determined target through multi-turn conversations. The key to achieving this task lies in planning dialogue paths that smoothly and coherently direct conversations towards the target. However, this is a challenging and under-explored task. In this work, we propose a coherent dialogue planning approach that uses a stochastic process to model the temporal dynamics of dialogue paths. We define a latent space that captures the coherence of goal-directed behavior using a Brownian bridge process, which allows us to incorporate user feedback flexibly in dialogue planning. Based on the derived latent trajectories, we generate dialogue paths explicitly using pre-trained language models. We finally employ these paths as natural language prompts to guide dialogue generation. Our experiments show that our approach generates more coherent utterances and achieves the goal with a higher success rate\footnote{Our code and data are available at \url{https://github.com/iwangjian/Color4Dial}.}.

\end{abstract}

\section{Introduction}
Dialogue systems have made significant progress in generating high-quality responses for open-domain chitchat \cite{zhang-etal-2020-dialogpt,roller-etal-2021-recipes} and assisting users in completing specific tasks \cite{madotto-etal-2018-mem2seq,wu2019global}. Instead of passively responding to users, dialogue systems can also take a proactive role to direct a conversation towards specific goals, such as introducing new and interesting topics \cite{wu-etal-2019-proactive} or providing sociable recommendations on target items \cite{wang2022follow}. Such a proactive target-oriented or goal-directed dialogue system can guide conversations towards topics that the system knows how to discuss, making it promising to build autonomous conversational AI.

\begin{figure}[th!]
\centering
\includegraphics[width=0.84\linewidth]{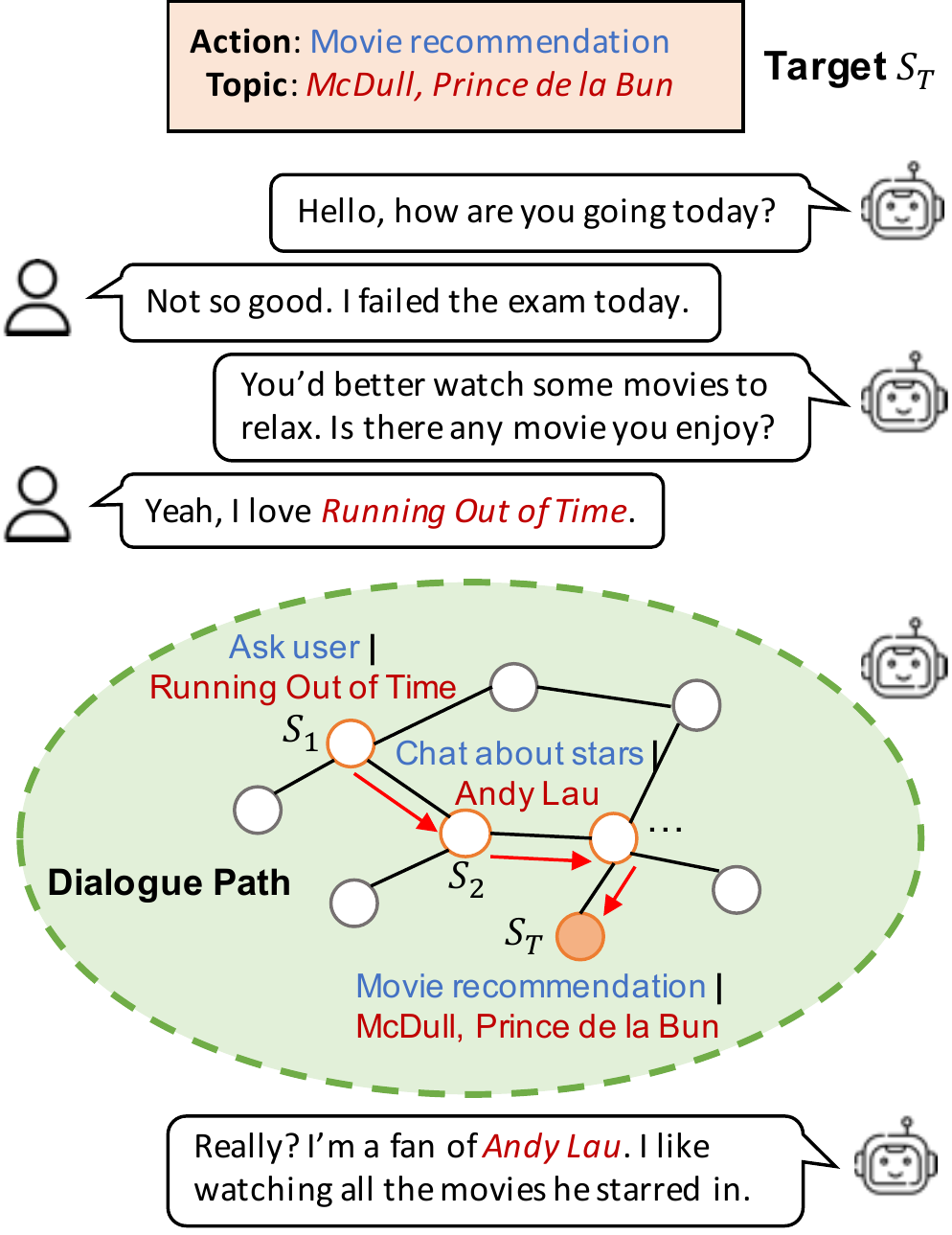}
\caption{An example from the repurposed DuRecDial 2.0 \cite{liu-etal-2021-durecdial} dataset. Given a pre-determined target and current dialogue context, we expect to plan a dialogue path to direct the conversation.}
\label{fig:example}
\end{figure}

For goal-directed dialogue systems, the objective is to proactively direct conversations towards a designated target. Previous work has primarily pre-determined the targets as specific keywords \cite{tang-etal-2019-target}, topics \cite{wu-etal-2019-proactive,sevegnani-etal-2021-otters}, and dialogue action-topic pairs \cite{zhang-etal-2021-kers-knowledge,wang2022follow}. To achieve this task, effective dialogue planning is essential, which requires taking reasonable actions and smoothly directing dialogue topics to the designated one. More importantly, the whole process is expected to be coherent and natural. 
Prior studies attempted to tackle this challenge through next-turn transition prediction \cite{tang-etal-2019-target}, sub-goal generation \cite{zhang-etal-2021-kers-knowledge,kishinami-etal-2022-target}, and knowledge path reasoning \cite{gupta-etal-2022-target} to control dialogue generation. However, there are still open issues worth exploring. 
\textbf{First}, previous studies adopted a greedy strategy with a single-turn topic prediction mechanism, which lacks global planning for the dialogue process \cite{yang-etal-2022-topkg}. Consequently, these methods are often short-sighted, resulting in sub-coherent topic threads.
\textbf{Second}, recognizing a user's engagement and willingness to follow the system is crucial for achieving coherent transitions. However, current studies often overlook the importance of modeling such user feedback. Therefore, it is necessary to explore globally planned dialogue strategies while incorporating user feedback to improve the coherence of goal-directed dialogue systems.

In this work, our objective is to globally plan dialogue paths that connect the current context to the target at each turn. As illustrated in Figure \ref{fig:example}, this dialogue path should strike a balance between coherence with the ongoing dialogue context and smooth transitions towards the target. Assuming that path trajectories without a target can be represented as Brownian motion \cite{revuz2013continuous} in latent space, we expect the embeddings of neighboring trajectory points to be similar to each other, while those of distant trajectory points to be dissimilar. Drawing inspiration from \citet{wang2022language}, we view goal-directed dialogue behavior as a Brownian bridge \cite{revuz2013continuous} stochastic process conditioned on fixed start and end points. As such, we can derive latent trajectories that follow coherent temporal dynamics.

Based on the above intuition, we propose a \textbf{\underline{c}}oherent dial\textbf{\underline{o}}gue p\textbf{\underline{l}}anning approach via Br\textbf{\underline{o}}wnian b\textbf{\underline{r}}idge (\textbf{\textsc{Color}}) stochastic process.
It involves mapping dialogue path points, such as topics or action-topic pairs, into a latent space of Brownian bridge conditioned on the current context and designated target. To ensure goal-directed behavior and incorporate user feedback, we also map the latest user utterance into real-time user feedback representation using the same latent space. We leverage this feedback to perturb the density and uncertainty of the Brownian bridge, simulating its impact on the dialogue planning process. Our training process uses a contrastive objective, which helps retain global coherence. We then fine-tune pre-trained language models (PLMs) using the derived latent trajectories to plan dialogue paths explicitly. These paths provide step-by-step explanations for reaching the target and serve as natural language prompts for generating system utterances.

In summary, our main contributions are: (1) We propose a novel approach called \textsc{Color}, which effectively models global coherence and incorporates user feedback in goal-directed dialogue planning. Our method utilizes the Brownian bridge stochastic process, and to the best of our knowledge, this is the first work to apply this method to the goal-directed proactive dialogue task. (2) We repurpose existing dialogue datasets by automatically constructing system goals and splitting them into in- and out-of-domain test sets. It facilitates research in the field and allows for more accurate evaluation of models. (3) Extensive experiments demonstrate that our proposed approach outperforms other methods, both in automatic and human evaluations.

\section{Preliminaries}

\paragraph{Problem Formulation}
We consider a corpus of goal-directed dialogues $\mathcal{D}=\{(\mathcal{K}_{i},\mathcal{P}_{i},\mathcal{C}_{i})\}_{i=1}^{N}$, where $N$ is the total number of dialogues. The domain knowledge facts relevant to the $i$-th dialogue are represented as $\mathcal{K}_{i}=\{k_{i,j}\}_{j=1}^{N_K}$, each in the form of a triple. The dialogue content for the $i$-th dialogue is $\mathcal{C}_{i}=\{\mathcal{C}_{i,t}\}_{t=1}^{N_T}$,  with a total of $N_T$ turns. The whole dialogue path for the $i$-th dialogue is denoted as $\mathcal{P}_{i}=\{\mathcal{P}_{i,l}\}_{l=1}^{L}$, where each path point is a topic or an action-topic pair. Here, dialogue topics are mainly constructed based on the domain knowledge $\mathcal{K}_{i}$. In some scenarios, there also exists a user profile $\mathcal{U}_{i}$, which can be user attributes or certain personal preferences.

Given a target $\mathcal{T}$ consisting of an action-topic pair or a topic only, a dialogue context $\mathcal{C}$, and a set of relevant domain knowledge $\mathcal{K}$ (and a user profile $\mathcal{U}$, if any), our objective is to generate coherent utterances to reach the target $\mathcal{T}$ when appropriate. The problem can be decomposed into two sub-tasks: (1) \textbf{dialogue planning}, which involves planning suitable actions and topics to lead the dialogue proactively with coherent transitions to the target, and (2) \textbf{dialogue generation}, which involves generating an appropriate utterance to achieve the planned action and topic at each turn.

\paragraph{Brownian Bridge}
The standard Wiener process or Brownian motion $W(t)$ has a normal distribution with mean $0$ and variance $t$, i.e., $W(t)\sim\mathcal{N}(0, t)$.
A Brownian bridge \cite{revuz2013continuous} is a continuous-time stochastic process pinned at fixed start and end points, where its distribution $B(t)$ is given by:
\begin{equation}
\small
    B(t)=W(t)-\frac{t}{T}W(T)
\end{equation}
where $t\in[0, T]$, $T$ denotes the end time. Furthermore, the transition distribution of a Brownian bridge process from an initial point $z_0$ at $t=0$ to an end point $z_T$ at $t=T$ is:
\begin{equation}
\small
    p(z_t|z_0, z_T)\sim\mathcal{N}\bigg(\Big(1-\frac{t}{T}\Big)z_{0}+\frac{t}{T}z_{T},\frac{t\big(T-t\big)}{T}\bigg)
\end{equation}
It implies that a trajectory point $z_t$ follows a noisy linear interpolation between $z_0$ and $z_T$, with $z_t$ closer to $z_0$ at the start and closer to $z_T$ at the end. The uncertainty is higher in the middle of the time interval and lower near the start and end points. The time-controlled nature of the Brownian bridge process has led to its application in various fields, such as trajectory simulation \cite{sousa2015brownian} and language modeling \cite{wang2022language}.

\section{Method}
We propose a \textbf{\underline{c}}oherent dial\textbf{\underline{o}}gue p\textbf{\underline{l}}anning approach via Br\textbf{\underline{o}}wnian b\textbf{\underline{r}}idge (\textbf{\textsc{Color}}) stochastic process to steer goal-directed dialogue generation. The intuition behind \textsc{Color} is to learn a mapping (see \S \ref{sec:sec1}) in the Brownian bridge latent space that captures coherent temporal dynamics for planning dialogue paths. Each dialogue path consists of a sequence of topics or action-topic pairs, starting from the current context and leading to the target. We generate these paths explicitly (see \S\ref{sec:sec2}) based on representations derived from the latent space, and use them to guide the generation of dialogue utterances (see \S\ref{sec:sec3}).

\subsection{Stage 1: Brownian Bridge Mapping}
\label{sec:sec1}

\begin{figure}[t!]
\centering
\includegraphics[width=0.94\linewidth]{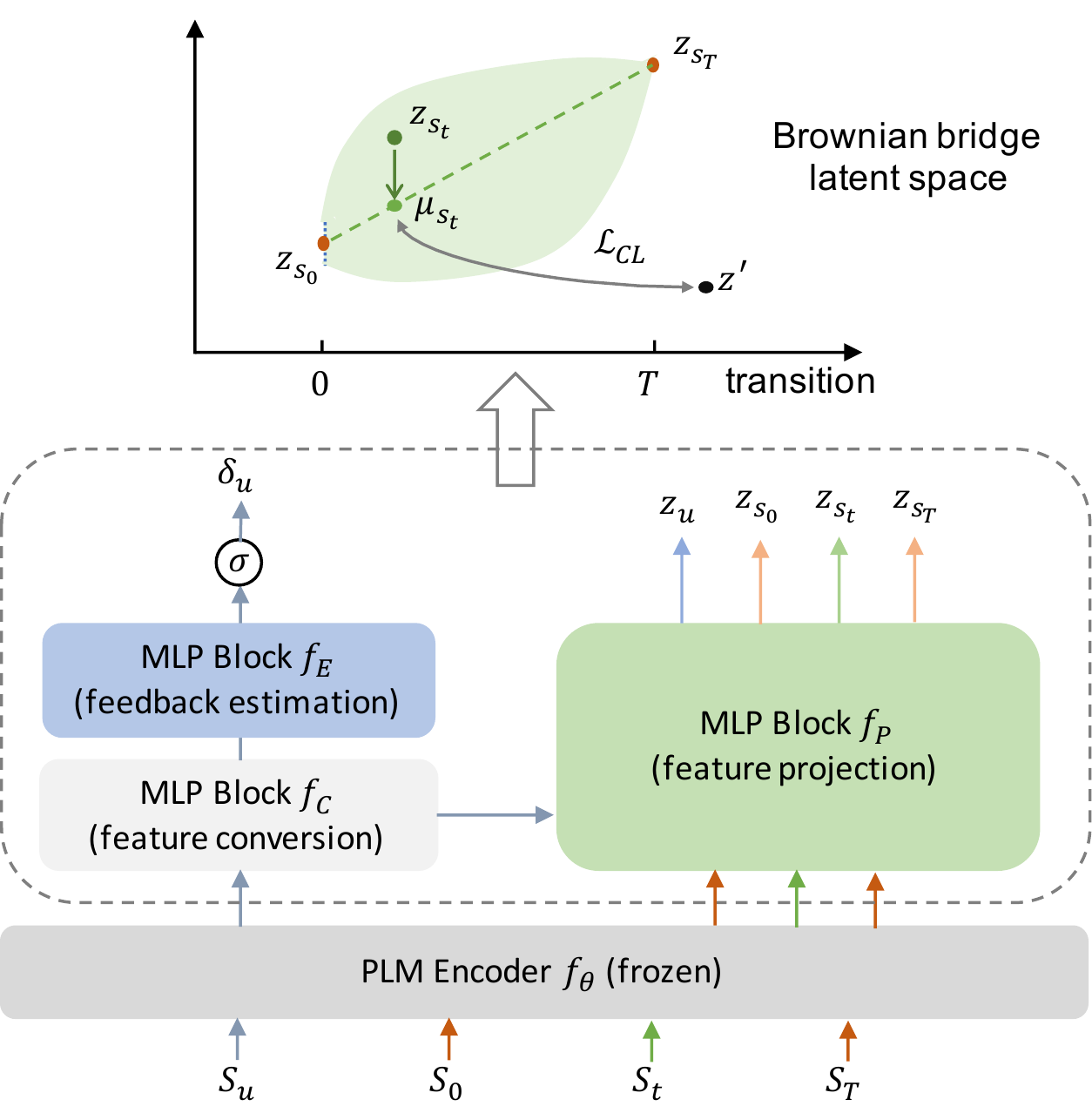}
\caption{Overview of Stage 1: Mapping observations to Brownian bridge latents. $S_u$ is the latest user utterance, $S_0$ is the concatenated text of domain knowledge and dialogue context, $S_T$ is the designated target, $S_t$ denotes a sampled path point in the dialogue path with $0<t<T$. We differentiate data flow by colored arrows.}
\label{fig:model_step1}
\end{figure}

A Brownian bridge latent space involves a non-linear mapping that transforms observations into a low-dimensional latent space, using the Brownian bridge stochastic process. Our objective is to utilize this mapping to train an encoder $\mathcal{F}$, to convert raw dialogue paths into latent representations that retain global coherence, with the overview depicted in Figure \ref{fig:model_step1}. In the following sections, we will introduce two crucial aspects of our approach: user feedback modeling and contrastive training.

\paragraph{User Feedback Modeling}
Suppose we obtain the user feedback representation $z_u$ and have an engagement indicator $\delta_{u}\in(0, 1)$, which reflects the user's level of engagement and likelihood of following the system, we newly define the transition distribution of the Brownian bridge process between a start point $z_{s_0}$ at $t=0$ and end point $z_{s_T}$ at $t=T$ as:
\begin{equation}
\resizebox{1.0\hsize}{!}{
    $p(z_{s_t})\sim\mathcal{N}\Big(\underbrace{\big(1-\frac{t}{T}\big)(z_{s_0}+z_{u})+\frac{t}{T}z_{s_T}}_{\mu_{s_t}}, \underbrace{\frac{t(T-t)}{T}+\varphi(\delta_{u})}_{\sigma^{2}}\Big)$
}
\label{eq:brownian}
\end{equation}
where $0<t<T$, $\varphi(\cdot)$ is a decaying function. Here, $z_u$ is used to perturb the density (the mean $\mu_{s_t}$) of the Brownian bridge process, and $\delta_{u}$ is used to perturb its uncertainty (the variance $\sigma^2$), with perturbation strength decaying over time. This decay means that the impact of the current user feedback on future planning is reduced.
Alternatively, $\varphi(\cdot)$ can be implemented with the linear decaying, i.e., $\varphi(\delta_{u})=\delta_{u}(1-t/T)$, or the exponential decaying, i.e., $\varphi(\delta_{u})=\delta_{u}e^{-t/(\lambda T)}$, where $\lambda\in(0,1)$ is a scaling factor.

\paragraph{Contrastive Training}
For a tuple of observations $(S_u, S_0, S_t, S_T)$, our objective is to ensure that their latent representations $(z_u, z_{s_0}, z_{s_t}, z_{s_T})$ follow the Brownian bridge transition distribution described in Eq. (\ref{eq:brownian}). Here, $S_u$ is the latest user utterance (and the concatenation of the user profile, if applicable), which may embody real-time user feedback information. $S_0$ consists of the concatenated domain knowledge and dialogue context, revealing the start of the dialogue path. $S_T$ is the designated target, representing the end of the dialogue path. A \textit{path point}, by default, refers to a topic or action-topic pair specific to the dataset. $S_t$ denotes a sampled path point in the dialogue path, $s.t.$, $0<t<T$. Here, $T$ denotes the number of transitions required to reach the target.

As shown in Figure \ref{fig:model_step1}, we build our encoder $\mathcal{F}$ on top of a frozen PLM encoder, which is followed by specific trainable multilayer perceptron (MLP) blocks. All the necessary latents are given by:
\begin{align}
\small
    z_{s_0} &= f_{P}\Big(\text{AvgPool}\big(f_{\theta}(S_0)\big)\Big), \\
    z_{s_t} &= f_{P}\Big(\text{AvgPool}\big(f_{\theta}(S_t)\big)\Big),  \\
    z_{s_T} &= f_{P}\Big(\text{AvgPool}\big(f_{\theta}(S_T)\big)\Big), \\
    z_u &= f_{P}\bigg(f_{C}\Big(\text{AvgPool}\big(f_{\theta}(S_u)\big)\Big)\bigg), \\
    \delta_{u} &= \sigma\bigg(f_{E}\Big(f_{C}\big(\text{AvgPool}(f_{\theta}(S_u))\big)\Big)\bigg)
\end{align}
where $f_{\theta}$ denotes a frozen PLM encoder such as BERT \cite{devlin-etal-2019-bert} or BART \cite{lewis-etal-2020-bart} encoder, AvgPool($\cdot$) denotes the average pooling operation. $f_P$, $f_C$, and $f_E$ are MLP blocks that produce output with a latent dimension of $d$. $\sigma$ is the Sigmoid activation function. The intuition behind the training is to ensure that the representation $z_{s_t}$ of a positive path point $S_t$ sampled from the same dialogue is close to the expected embedding $\mu_{s_t}$ (the mean in Eq. (\ref{eq:brownian})). In contrast, the representation $z^{'}$ of a negative random path point $S_{t}^{'}$ from a different dialogue is far from $\mu_{s_t}$ (see Figure \ref{fig:model_step1}) because it does not align with the Brownian bridge pinned by $z_{s_0}$ and $z_{s_T}$.
We consider a contrastive objective proposed in \citet{wang2022language} for training. Formally, given input batches $\mathcal{B}=\{(S_u, S_0, S_t, S_T)\}$ consisting of randomly sampled positive path points $S_t$ where $0<t<T$, we optimize our encoder $\mathcal{F}$ as follows:
\begin{align}
\small
    \mathcal{L}_{CL} &= -\log \frac{\exp(\text{d}(S_{t}^{+}; \mathcal{F}))}{\sum\limits_{S_{t}^{-} \in \mathcal{B}} \exp(\text{d}(S_{t}^{-}; \mathcal{F}))}, \label{eq:cont_loss} \\ 
    \text{d}(S_t; \mathcal{F}) &= -\frac{1}{2\sigma^2}\bigg\|z_{s_t}-\mu_{s_t}\bigg\|^2_2 \label{eq:distance_func}
\end{align}
where $S_{t}^{+}$ denotes a positive tuple $(S_u, S_0, S_t, S_T)$, $S_{t}^{-}$ denotes a negative tuple $(S_u, S_0, S_{t}^{'}, S_T)$, $\sigma^2$ is the variance in Eq. (\ref{eq:brownian}), $\mu_{s_t}$ is the mean in Eq. (\ref{eq:brownian}).

\subsection{Stage 2: Planning Dialogue Paths}
\label{sec:sec2}

The Brownian bridge latent space makes it easy to derive a coherent latent trajectory with temporal dynamics. We feed the start point $S_0$, designated target $S_T$, and observed $S_u$, into the trained encoder $\mathcal{F}$ respectively, then sample a latent trajectory $z=(z_{s_1},z_{s_2},\cdots,z_{s_T})$ that follows Eq. (\ref{eq:brownian}), where $z_{s_t}\in\mathbb{R}^{d}$, $t=1,2,\cdots,T$. Here, $z$ acts like the transition-level latent representation that connects the ongoing dialogue context to the target, i.e., the dialogue path $\mathcal{P}$ to be planned. 

\begin{figure}[t!]
\centering
\includegraphics[width=0.94\linewidth]{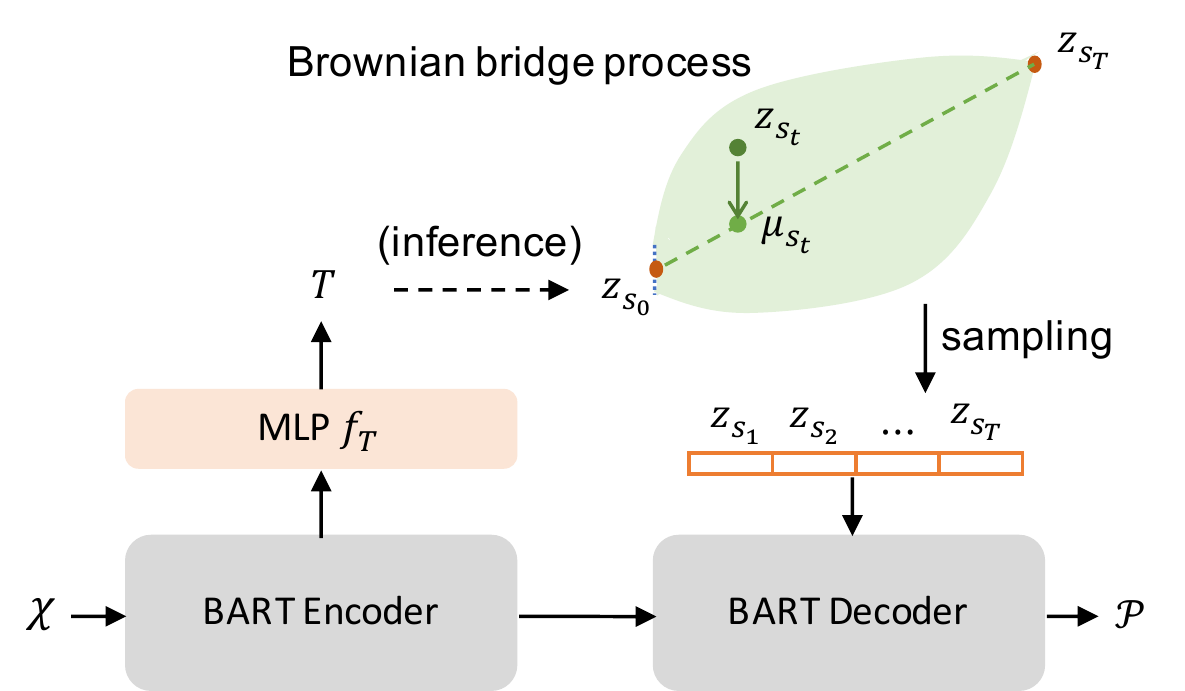}
\caption{Overview of Stage 2: Planning the dialogue path $\mathcal{P}$, where $\mathcal{X}$ is the required input, $T$ denotes the number of transitions required to reach the target.}
\label{fig:model_step2}
\end{figure}

To generate the path $\mathcal{P}$, we define the required input as $\mathcal{X}=[\mathcal{C};\mathcal{K};\mathcal{T}]$, which is the concatenated text of the dialogue context $\mathcal{C}$, domain knowledge $\mathcal{K}$, and target $\mathcal{T}$. As shown in Figure \ref{fig:model_step2}, we feed $\mathcal{X}$ into a pre-trained BART \cite{lewis-etal-2020-bart} model for fine-tuning, with the encoded hidden states being $h=(h_1, h_2, \cdots, h_m)$. We discuss the generation of $\mathcal{P}$ by conditioning on $h$ and $z$ below.

\textbf{First}, sampling the latent trajectory $z$ requires the value $T$, i.e., the number of transitions to reach the target. We obtain this value by adding an MLP layer $f_T$ to the BART encoder as a predictor, which outputs the probability of $T$:
\begin{equation}
    p(T)=\text{softmax}(W_{1}f_{T}(\bar{h})+b_{1})
\label{eq:pred_t}
\end{equation}
where $\bar{h}$ is the average pooled representation of $h$, $W_1$ and $b_1$ are trainable parameters. We optimize the predictor using cross-entropy loss $\mathcal{L}_{c}$.

\textbf{Second}, our BART decoder conditions on $h$ and the derived latent trajectory $z$, then generates the dialogue path $\mathcal{P}$ with encoder-decoder attentions. The output distribution is approximated as follows:
\begin{align}
    p_{\theta}(\hat{y}_t)&=\text{softmax}(W_{2}h_{t}^{o}+b_2), \\
    h_t^{o}&=\text{Decoder}(y_{t-1}, H), \label{eq:dec_hidden} \\
    H&=[h;W^{\mathrm{T}}z] \label{eq:z_hidden}
\end{align}
where $W_2$, $b_2$ are trainable parameters, $W$ denotes a linear transformation that maps the dimension of $z$ to be identical to $h$, and $[;]$ denotes concatenation. The decoder is trained by minimizing the negative log-likelihood below:
\begin{equation}
    \mathcal{L}_{g}= -\sum_{i=1}^{N}p({y}^{(i)})\log p_{\theta}(\hat{y}^{(i)})
\end{equation}
where $p(y^{(i)})$ is the distribution of the ground-truth dialogue path, while $p_{\theta}(\hat{y}^{(i)})$ is the distribution of the approximated dialogue path.

In addition, for the decoder's all hidden states $h^{o}=(h_{1}^{o},h_{2}^{o},\cdots,h_{n}^{o})$ (see Eq. (\ref{eq:dec_hidden})) and the transformed latent trajectory $z^{o}=W^{\mathrm{T}}z$ (see Eq. (\ref{eq:z_hidden})), they inevitably both represent the dialogue path $\mathcal{P}$ though at different levels. We minimize the Kullback–Leibler (KL) divergence between $h^{o}$ and $z^{o}$:
\begin{equation}
    \mathcal{L}_{KL}=\sum_{i=1}^{N}D_{KL}({\bar{h^o}}^{(i)}||{\bar{z^o}}^{(i)})
\end{equation}
where $\bar{h^o}$ and $\bar{z^o}$ denote the average pooled representation of $h^{o}$ and $z^{o}$, respectively. 

For training, our model is optimized as follows:
\begin{equation}
\mathcal{L}=\alpha\mathcal{L}_{c}+\beta\mathcal{L}_{g}+\gamma\mathcal{L}_{KL}
\end{equation}
where $\alpha$, $\beta$, and $\gamma$ are hyperparameters. During inference, we obtain the value $T$ inferred by the predictor $f_T$, then sample a latent trajectory $z=(z_{s_1},\cdots,z_{s_T})$ given $t=1,\cdots,T$. The decoder then generates a dialogue path token by token. Additionally, no transition is needed to reach the target if $T=0$. In such cases, we directly generate the dialogue path by copying the given target $\mathcal{T}$.

\subsection{Stage 3: Generating Dialogue Utterances}
\label{sec:sec3}

Motivated by prior work on prompt-based learning for dialogue generation \cite{zheng2021exploring,madotto2021few}, we regard each dialogue path $\mathcal{P}$ as a natural language prompt to guide a generative PLM for dialogue generation. Here, $\mathcal{P}$ serves as a global prompt that outlines the dialogue actions and topics needed to reach the target step by step. With the power of the PLM, $\mathcal{P}$ helps to distill the necessary knowledge from both the input text and the PLM.
To formulate the newly input $\mathcal{X}^{'}$, we append $\mathcal{P}$ to the given dialogue context $\mathcal{C}$ and domain knowledge $\mathcal{K}$, and concatenate them as:
\begin{equation}
  \mathcal{X}^{'}=[\mathcal{K};\mathcal{C};\mathcal{P}]
\end{equation}
where $[;]$ denotes concatenation. We then feed $\mathcal{X}^{'}$ into a pre-trained GPT-2 \cite{radford2019language} or DialoGPT \cite{zhang-etal-2020-dialogpt} for supervised fine-tuning. We adopt the planned dialogue paths generated by our \textsc{Color} during inference.

\begin{table}[t!]
\centering
\resizebox{0.96\linewidth}{!}{
\begin{tabular}{llcccc}
\toprule
 \multicolumn{2}{c}{\multirowcell{2}{\textbf{Dataset}}}  & \multirow{2}{*}{\textbf{\#Dial.}} & \multirow{2}{*}{\textbf{\#Utter.}} & \multicolumn{2}{c}{\textbf{Dial. Turn}}  \\
 &  &  &   &  \textbf{\#Max.} &  \textbf{\#Avg.}  \\
\midrule
 \multirow{4}{*}{DuRecDial 2.0} & Train  & 4,256  & 68,781  & 13  & 8.1   \\
 & Valid   &  608 &  9,677  & 14  & 8.0 \\
 & Test-ID  &  770 &  12,299 & 13  & 8.0 \\
 & Test-OOD  &  446 &  7,962 & 12  & 8.9  \\
 \midrule
 \multirow{3}{*}{TGConv} & Train  & 15,197  & 70,205  & 9  & 3.8 \\
  &  Valid  & 2,681  & 12,167   & 9  & 3.7  \\
  &  Test  & 1,000  &  5,132  &  9 &  3.9  \\
\bottomrule
\end{tabular}}
\caption{Overview of the datasets.}
\label{tab:data_stats}
\end{table}

\section{Experiments and Results}

\subsection{Experimental Setup}

\paragraph{Datasets}
The task of goal-directed proactive dialogue is under-explored, making it challenging to find feasible benchmarks for evaluation. After careful consideration, we have identified the DuRecDial 2.0 \cite{liu-etal-2021-durecdial} and TGConv \cite{yang-etal-2022-topkg} datasets as appropriate for our experiments. 
DuRecDial 2.0 \cite{liu-etal-2021-durecdial} is a crowdsourced dataset of human-to-human dialogues in recommendation-oriented scenarios. We repurpose the dataset by defining the targets as action-topic pairs. We obtain two types of splits for the test set: \textit{in-domain} (ID) and \textit{out-of-domain} (OOD), similar to \citet{sevegnani-etal-2021-otters}. The OOD split ensures that none of the target topics in the test set are present in the training set, whereas the ID split allows them to appear. 
The TGConv \cite{yang-etal-2022-topkg} dataset contains high-quality open-domain dialogues on a variety of commonsense topics. Each dialogue is designed to direct the conversation towards a specific keyword or topic through coherent keyword transitions, which are categorized as either easy-to-reach or hard-to-reach based on their difficulty level. Table \ref{tab:data_stats} summarizes the statistics of both datasets. More details are available in Appendix \ref{appendix:dataset}.

\begin{table*}[t!]
\centering
\resizebox{0.9\textwidth}{!}{
\begin{tabular}{clcccccc}
\toprule
\textbf{Split} & \textbf{Model} & \textbf{PPL ($\downarrow$)} & \textbf{F1 (\%)}  & \textbf{BLEU-1 / 2}  & \textbf{DIST-1 / 2}  & \textbf{Know. F1 (\%)} & \textbf{Succ. (\%)} \\
\midrule 
\multirow{8}{*}{ID} & MGCG\_G \cite{liu-etal-2020-towards-conversational} & 25.32  & 35.13 & 0.316 / 0.211  & 0.016 / 0.053  &  39.53 & 29.49  \\
  & KERS \cite{zhang-etal-2021-kers-knowledge} & 20.15  &  31.27 & 0.288 / 0.196 & 0.017 / 0.061  & 41.18  &  33.75 \\
  &  GPT-2 \cite{radford2019language} & 5.33 &  36.86 &  0.314 / 0.222 & 0.024 / 0.081 & 43.62 & 41.80 \\
  &  DialoGPT \cite{zhang-etal-2020-dialogpt} & 5.26  & 38.12 & 0.324 / 0.252 & 0.023 / 0.076  & 44.71  & 46.46 \\
  &  BART \cite{lewis-etal-2020-bart} &  6.46 & 36.11  & 0.279 / 0.181 & \textbf{0.030} / \textbf{0.096} & 43.33  & 58.40   \\
  & TCP-Dial \cite{wang2022follow} &  5.88 &  34.46 &  0.293 / 0.201  & 0.027 / 0.091  & 45.75 & 60.49  \\
\cmidrule{2-8}
 & Ours (\textsc{Color} w/ GPT-2) &  \textbf{5.17} &  40.43* & 0.337* / 0.243* & 0.026 / 0.084  & 50.81*  & 69.14*  \\
 & Ours (\textsc{Color} w/ DialoGPT) & 5.22 & \textbf{43.14}*  & \textbf{0.371}* / \textbf{0.277}* & 0.024 / 0.073  & \textbf{57.89}* &  \textbf{73.20}*  \\
\midrule
\multirow{8}{*}{OOD} & MGCG\_G \cite{liu-etal-2020-towards-conversational} &  28.21 & 30.84 & 0.276 / 0.167  &  0.015 / 0.046  &  20.53 & 8.46 \\
  & KERS \cite{zhang-etal-2021-kers-knowledge}  & 24.35  &  27.91  & 0.259 / 0.160  &  0.016 / 0.058  & 26.88  & 14.15  \\
  &  GPT-2 \cite{radford2019language} & 5.86 &  33.06 & 0.276 / 0.193  & 0.023 / 0.077  & 28.79  &  32.79 \\
  &  DialoGPT \cite{zhang-etal-2020-dialogpt} &  5.37 & 34.27  & 0.283 / 0.176  &  0.021 / 0.068 & 31.75  & 32.47 \\
  &  BART \cite{lewis-etal-2020-bart} & 8.09 & 32.38  & 0.244 / 0.149 & 0.026 / 0.081  & 30.02 & 43.08  \\
  & TCP-Dial \cite{wang2022follow} &  8.24 & 29.24  & 0.255 / 0.165 & \textbf{0.027} / \textbf{0.089}  & 21.36 & 18.40 \\
\cmidrule{2-8}
 & Ours (\textsc{Color} w/ GPT-2) & 5.63 &  34.44*  &  0.285* / 0.198*  &  0.025 / 0.082  &  34.04* &  \textbf{57.41}*  \\
 & Ours (\textsc{Color} w/ DialoGPT) & \textbf{5.30} & \textbf{37.97}*  & \textbf{0.320}* / \textbf{0.227}*   &  0.024 / 0.072  & \textbf{41.35}*  & 52.36* \\
\bottomrule
\end{tabular}}
\caption{Automatic local and global evaluation results of dialogue generation on the DuRecDial 2.0 dataset with different test splits. Significant improvements over backbone models are marked with * (t-test, $p < 0.05$).}
\label{tab:dialogue_durecdial}
\end{table*}

\paragraph{Baseline Methods}
For dialogue generation, our baselines include: \textbf{GPT-2} \cite{radford2019language}, \textbf{DialoGPT} \cite{zhang-etal-2020-dialogpt}, and \textbf{BART} \cite{lewis-etal-2020-bart}. On the repurposed DuRecDial 2.0 dataset, we also compared our method with three competitive methods: \textbf{MGCG\_G} \cite{liu-etal-2020-towards-conversational}, \textbf{KERS} \cite{zhang-etal-2021-kers-knowledge}, and \textbf{TCP-Dial} \cite{wang2022follow}. We chose these methods because they are highly relevant to our problem setting, and TCP-Dial is currently the state-of-the-art model in our knowledge.
Given that our method is generalizable to the existing TGConv dataset, we evaluated its effectiveness against four competitive models specific to that dataset: \textbf{MultiGen} \cite{ji-etal-2020-language}, \textbf{DKRN} \cite{qin2020dynamic}, \textbf{CKC} \cite{zhong2021keyword}, and \textbf{TopKG} \cite{yang-etal-2022-topkg}. More details about the above methods are shown in Appendix~\ref{appendix:baseline_gen}. 
For dialogue planning, we compared our \textsc{Color} with the planning models proposed in the above methods using a planning-enhanced paradigm. We also included \textbf{BERT} \cite{devlin-etal-2019-bert} and \textbf{BART} \cite{lewis-etal-2020-bart} as our baselines. More details about them are described in Appendix~\ref{appendix:baseline_plan}.

\paragraph{Implementation Details}
\label{sec:implement}
Our proposed \textsc{Color} model is implemented by PyTorch. In both Stage 1 and Stage 2, we adopt the BART-base model (768 dimensions, 6 encoder/decoder layers, and 12 attention heads) released in Huggingface's Transformers \cite{wolf-etal-2020-transformers} library. The latent dimension $d$ is set to 16. The MLP blocks $f_P$, $f_C$, and $f_E$ are all stacked to 3 layers. The decaying function $\varphi(\cdot)$ employs the linear decaying. The hyperparameters $\alpha$, $\beta$ and $\gamma$ are set to 0.1, 1.0 and 1.0, respectively. For training in Stage 2, we construct the dialogue path $\mathcal{P}$ in the format of $\texttt{[A]}a_1\texttt{[T]}t_1\cdots\texttt{[A]}a_T\texttt{[T]}t_T$ on the DuRecDial 2.0, and of $\texttt{[T]}t_1\cdots\texttt{[T]}t_T$ on the TGConv. Here, \texttt{[A]} is a special token to separate an action $a_{i}$, \texttt{[T]} is a special token to separate a topic $t_{i}$. During inference, we generate a dialogue path token by token. Further details on training and inference are provided in Appendix~\ref{appendix:training}.

\subsection{Evaluation of Dialogue Generation}

\begin{table}[t!]
\centering
\resizebox{1.0\linewidth}{!}{
\begin{tabular}{lcccc}
\toprule
\multirowcell{2}{\textbf{Model}} &  \multicolumn{2}{c}{\textbf{Easy Target}}  & \multicolumn{2}{c}{\textbf{Hard Target}} \\
 & \textbf{Succ. (\%)}  & \textbf{Coh.}  &  \textbf{Succ. (\%)}  & \textbf{Coh.} \\
\midrule
GPT-2\textsuperscript{$\dagger$} (G) & 22.3 & 0.23 & 17.3  & 0.21 \\
DialoGPT (D) & 32.3 & 0.30 & 23.8  & 0.25  \\
MultiGen\textsuperscript{$\dagger$} & 26.7 &  0.21 & 19.6 & 0.24 \\
DKRN\textsuperscript{$\dagger$}  & 38.6  &  0.33 & 21.7 & 0.31  \\
CKC\textsuperscript{$\dagger$}  &  41.9  & 0.35  & 24.8 & 0.33  \\
TopKG\textsuperscript{$\dagger$}  & 48.9  &  0.31 & 27.3  & 0.33  \\
\midrule
Ours (\textsc{Color} w/ G) & 54.2  & 0.34  & 28.8 & 0.33  \\
Ours (\textsc{Color} w/ D) & \textbf{66.3}  & \textbf{0.36}  & \textbf{30.1} & \textbf{0.35} \\
\bottomrule
\end{tabular}}
\caption{Automatic global evaluation results of dialogue generation on the TGConv dataset. G and D are short for GPT-2 and DialoGPT, respectively. Models marked with $\dagger$ are reported from \citet{yang-etal-2022-topkg}.}
\label{tab:dialogue_tgconv}
\end{table}

\paragraph{Evaluation Metrics}
To evaluate the performance of next-turn system utterance generation, we adopt commonly used local evaluation metrics, including perplexity (\textbf{PPL}), distinct (\textbf{DIST}) \cite{li-etal-2016-diversity}, \textbf{BLEU}-$n$ \cite{papineni-etal-2002-bleu}, word-level \textbf{F1} and knowledge F1 (\textbf{Know. F1}) \cite{liu-etal-2020-towards-conversational}. 
To evaluate models' goal-directed performance, we use the goal success rate (\textbf{Succ.}) as the global evaluation metric. In the repurposed DuRecDial 2.0 dataset, Succ. measures the proportion of correct target topic generation within the target turn and the two adjacent turns in the test set, as per \citet{wang2022follow}. For the TGConv dataset, we perform self-play simulations, following \citet{yang-etal-2022-topkg}, to simulate multi-turn conversations and compute the success rate of generating the target keyword within 8 turns. Additionally, we adopt coherence (\textbf{Coh.}) \cite{yang-etal-2022-topkg} as another global evaluation metric, which measures the average contextual semantic similarity between the last utterance in the context and generated utterance.

\paragraph{Results and Discussion}
Table \ref{tab:dialogue_durecdial} shows evaluation results on the DuRecDial 2.0 dataset. We observe that MGCG\_G and KERS achieve comparable performance to PLM-based models on the in-domain (ID) split. One main reason is that they use the predicted dialogue action and topic to guide the model in utterance generation. However, they perform poorly in terms of goal success rate due to a lack of dialogue-level planning. We note that BART and TCP-Dial obtain much better DIST-1/2 scores than others because they seldom generate repeated words, making the generated utterances more diverse. In comparison, our models achieve remarkable improvements over most evaluation metrics. For example, our \textsc{Color} with DialoGPT achieves much better knowledge F1 scores, indicating that our method is more likely to generate utterances with correct knowledge. Regarding the goal success rate, our models obtain a large margin of improvement on both ID and OOD splits. It shows that using prompts with appropriate dialogue paths effectively steers PLMs to generate proper utterances for goal-directed dialogue. 

As shown in Table \ref{tab:dialogue_tgconv}, we notice that directing a dialogue to reach the target seems challenging in the context of open-domain chitchat for all models. However, with the guidance of our dialogue planning approach, \textsc{Color}, our models are able to produce more coherent utterances and reach the target at a significantly higher success rate.

\subsection{Evaluation of Dialogue Planning}

\paragraph{Evaluation Metrics}
To evaluate the performance of dialogue planning, we first adopt \textbf{F1} to measure the micro-averaged precision and recall of the predicted action or topic. For generation-based models, we extract the action or topic at the evaluated turn from the generated dialogue path for a fair comparison. Due to the nature of dialogue, multiple temporary planning strategies can be reasonable before reaching the target. Following \citet{zhou2020augmenting}, we also expand gold labels by considering the system's actions or topics within the previous and subsequent turns. As such, we then compute bigram action F1 (\textbf{Bi-act. F1}) and bigram topic F1 (\textbf{Bi-top. F1}) for evaluation.

\begin{table}[th!]
\centering
\resizebox{1.0\linewidth}{!}{
\begin{tabular}{clcccc}
\toprule
\multirow{2}{*}{\textbf{Split}} &  \multirow{2}{*}{\textbf{Model}} & \multicolumn{2}{c}{\textbf{Action}} &  \multicolumn{2}{c}{\textbf{Topic}} \\
 &  & \textbf{F1} &  \textbf{Bi-act. F1}  & \textbf{F1} & \textbf{Bi-top. F1}  \\
\midrule
\multirow{6}{*}{ID} & MGCG & 90.26 &  92.47 &  74.93  & 79.24  \\
 &  KERS & 90.33 &  91.54 &  77.85  & 80.35 \\
 &  BERT & 91.68 &  92.37 &  80.64  & 82.59 \\
 &  TCP  & 92.25 &  93.82 &  85.77  & 87.25 \\
 &  BART & 95.40 &  96.31 &  90.96  & 92.21 \\ 
\cmidrule{2-6}
 & Ours (\textsc{Color}) & \textbf{96.86} & \textbf{97.68}  & \textbf{93.30}  & \textbf{94.26} \\
\midrule
\multirow{6}{*}{OOD} & MGCG & 82.30 & 87.25  & 36.03  & 42.00\\
 & KERS & 84.21 & 86.39  & 34.20  & 37.85 \\
 & BERT & 92.23 & 94.19  & 46.55  & 52.12 \\
 & TCP  & 89.93 & 92.09  & 44.49  & 50.71 \\
 & BART & 92.63 & 93.18  & 58.57  & 62.37 \\ 
\cmidrule{2-6}
 & Ours (\textsc{Color}) & \textbf{93.43} & \textbf{93.82}  & \textbf{79.09}  & \textbf{83.46} \\
\bottomrule
\end{tabular}}
\caption{Results of dialogue planning on the DuRecDial 2.0 with different test splits.}
\label{tab:planning_durecdial}
\end{table}

\begin{table}[th!]
\centering
\resizebox{0.88\linewidth}{!}{
\begin{tabular}{lcc}
\toprule
\textbf{Model} & \textbf{F1}  & \textbf{Bi-top. F1} \\
\midrule
BERT \cite{devlin-etal-2019-bert} & 45.90 & 49.17 \\
BART \cite{lewis-etal-2020-bart} & 43.20 & 47.69 \\
TopKG-Plan \cite{yang-etal-2022-topkg} &  46.06 & 48.04 \\
\midrule
Ours (\textsc{Color}) & \textbf{47.17}  & \textbf{52.85}  \\
\bottomrule
\end{tabular}}
\caption{Results of dialogue planning on the TGConv.}
\label{tab:planning_tgconv}
\end{table}

\paragraph{Results and Discussion}
Table \ref{tab:planning_durecdial} reports the evaluation results on the DuRecDial 2.0 dataset. We find that predicting or generating dialogue topics is more challenging than dialogue actions. Further analysis reveals that the dialogue actions follow a similar transition pattern in the dialogue paths, making it easier for all models to predict actions with an F1 score of over 80\%. On the other hand, the variation in dialogue paths is primarily related to topics, which requires complex reasoning of domain knowledge, dialogue context, and target  for accurate prediction.
When evaluating on the OOD split, all baselines show lower F1 and Bi-top. F1 scores for topics. However, our proposed \textsc{Color} achieves substantial improvements. We observe similar trends in Table \ref{tab:planning_tgconv} when evaluating on the TGConv dataset. Overall, our \textsc{Color} outperforms the baselines by generating more reasonable actions and appropriate topics, making it a promising approach for planning dialogue paths.

\paragraph{Analysis of Model Variants}
We analyze the following variants of our model: (1) $\textbf{\textsc{Color}}_{d=?}$, which varies the value of the latent dimension $d$ in $\{8, 32, 128\}$ (The $d$ in our \textsc{Color} is set to 16 as described in \S \ref{sec:implement}); (2) \textit{\textbf{w/o}} Brownian bridge (\textbf{BB}), which removes the operation of conditioning on the derived Brownian bridge latent representation $z$; (3) \textit{\textbf{w/o}} user feedback modeling (\textbf{UFM}), which removes $z_u$ and $\varphi(\delta_u)$ in our Brownian bridge process as defined in Eq. (\ref{eq:brownian}); (4) \textit{\textbf{w/o}} $\mathcal{L}_{KL}$, which means the model is trained without the loss $\mathcal{L}_{KL}$.

\begin{table}[th!]
\centering
\resizebox{0.92\linewidth}{!}{
\begin{tabular}{lcccc}
\toprule
\multirow{2}{*}{\textbf{Model}} & \multicolumn{2}{c}{\textbf{Action}} &  \multicolumn{2}{c}{\textbf{Topic}} \\
 & \textbf{F1} &  \textbf{Bi-act. F1}  & \textbf{F1} & \textbf{Bi-top. F1}  \\
\midrule
$\textsc{Color}_{d=8}$ & 93.21  & 93.73  & 79.21  &  83.30 \\
$\textsc{Color}_{d=32}$ & 91.24  & 92.82  & 78.03   & 83.34 \\
$\textsc{Color}_{d=128}$ & 93.57  & 94.30  & 78.67  & 82.89 \\
\midrule
\textsc{Color} & 93.43 & 93.82 & 79.09  & 83.46  \\
\quad\textit{w/o} BB  &  93.66  &  93.93  &   62.45 & 64.27  \\
\quad\textit{w/o} UFM  &  92.42  &  92.84  & 77.21 & 80.57 \\
\quad\textit{w/o} $\mathcal{L}_{KL}$  & 92.95  &  93.01 &  77.34  &  80.97 \\ 
\bottomrule
\end{tabular}}
\caption{Dialogue planning performance of our \textsc{Color} with different variants.}
\label{tab:ablation}
\end{table}

We report evaluation results on the OOD split of the DuRecDial 2.0 dataset, as shown in Table~\ref{tab:ablation}. We observe that a larger value of $d$ brings fewer performance gains. Hence, the $d$ in our \textsc{Color} is set to 16 after making a trade-off between effectiveness and efficiency. We note that each module or mechanism of \textsc{Color} contributes to dialogue planning. In particular, the performance of \textsc{Color} sharply drops without the Brownian bridge (BB). It is because the derived Brownian bridge latent trajectory serves as a transition-level latent representation of the dialogue path to be planned. More importantly, it follows coherent temporal dynamics and thus benefits planning the dialogue path.

\subsection{Human Evaluation}

We recruit three well-educated graduate students as annotators for human evaluation. We ask the annotators to score different models based on turn-level and dialogue-level metrics, following \citet{liu-etal-2020-towards-conversational}. The turn-level evaluation measures appropriateness (\textbf{Appr.}) and informativeness (\textbf{Info.}). The dialogue-level evaluation measures proactivity (\textbf{Proact.}), coherence (\textbf{Coh.}), and goal success (\textbf{Succ.}). More details on the metrics and evaluation procedure are described in Appendix~\ref{appendix:humaneval}.

Table \ref{tab:human_eval} shows human evaluation results on the DuRecDial 2.0 dataset. The Fleiss's kappa \cite{fleiss1971measuring} scores are mainly distributed between [0.41, 0.60], indicating moderate inter-annotator agreement. We observe that DialoGPT, TCP-Dial, and ours obtain comparable scores in informativeness since they all utilize powerful PLMs. However, our method is able to generate more appropriate utterances in response to dialogue context. For dialogue-level evaluation, our method obtains better results on average compared to all baseline models. Notably, our method achieves the highest coherence score and goal success rate, indicating that our method is more likely to direct the dialogue to reach the target coherently and successfully.

\begin{table}[t!]
\centering
\resizebox{0.88\linewidth}{!}{
\begin{tabular}{lccccc}
\toprule
\textbf{Model} & \textbf{Appr.}  & \textbf{Info.} & \textbf{Proact.}  & \textbf{Coh.} & \textbf{Succ.} \\
\midrule
MGCG\_G  & 0.84 & 1.02 & 0.92 & 0.92 & 0.90 \\
DialoGPT & 1.17 & 1.35 & 1.06 & 1.17 & 1.19\\
TCP-Dial & 1.20 & 1.24 & 1.26 & 1.20 & 1.02 \\
Ours  & \textbf{1.33}  & \textbf{1.40}  & \textbf{1.42}  & \textbf{1.35} & \textbf{1.38} \\
\hline
kappa & 0.48  & 0.52  & 0.46  & 0.56  & 0.53\\
\bottomrule
\end{tabular}}
\caption{Human evaluation results. The Fleiss’s kappa measures the agreement among the annotators.}
\label{tab:human_eval}
\end{table}

\subsection{Case Study}
To better analyze goal-directed dialogue generation, we show some cherry-picked cases in Appendix \ref{appendix:case_study} due to space limitation. We observe that some baseline models can generate fluent and informative utterances. However, they still fail to direct the dialogue to reach the target and are ineffective to maintain coherence.
In comparison, our \textsc{Color} model can plan a dialogue path with reasonable actions and appropriate topics that outlines how to reach the target step by step. With the guidance of the planned dialogue path, our system better knows when and what to talk about to proactively move the dialogue forward. More importantly, our method succeeds in achieving the goal (see Appendix \ref{appendix:case_study}).

\section{Related Work}
\label{sec:relwork}

\paragraph{Goal-directed Dialogue Generation}
In the goal-directed or target-oriented setting, existing studies mainly predetermine the targets as specific keywords \cite{tang-etal-2019-target,qin2020dynamic,zhong2021keyword}, topics \cite{wu-etal-2019-proactive,sevegnani-etal-2021-otters,lei2022interacting}, and dialogue action-topic pairs \cite{zhang-etal-2021-kers-knowledge, wang2022follow}. The key to the task is dialogue planning, which leads the dialogue towards the target smoothly and coherently. Prior work pays attention to next-turn transition strategy \cite{tang-etal-2019-target}, hierarchical policy \cite{xu-etal-2020-conversational,xu2020knowledge}, and sub-goal generation \cite{zhang-etal-2021-kers-knowledge, kishinami-etal-2022-target}. For this knowledge-rich task, recent work \cite{gupta-etal-2022-target,yang-etal-2022-topkg,wang2022follow} further concerns planning a dialogue path based on grounded knowledge to guide every turn of response generation.

\paragraph{Planning for Language Generation}
There is a line of work \cite{puduppully2019data,hua-wang-2019-sentence,moryossef-etal-2019-step,su-etal-2021-plan-generate} that separates text generation into content planning and surface realization. Content planning mainly concerns selecting key content (e.g., key entities) and arranging their orders. Several planning frameworks \cite{hua-etal-2021-dyploc,hu-etal-2022-planet,li-etal-2022-event} have been studied to control complex language generation tasks. Our work is more related to planning for dialogue generation \cite{kishinami-etal-2022-target,yang-etal-2022-topkg,cohen2022dynamic}. Our proposed \textsc{Color} is a novel dialogue-level planning method that steers dialogue generation.

\section{Conclusion}
In this work, we explore the task of goal-directed proactive dialogue and focus on planning dialogue paths that direct conversations towards the designated target. We propose a novel approach called \textsc{Color}, which models coherent temporal dynamics for dialogue paths in the defined latent space, and considers the impact of user feedback on the dialogue planning process. We employ the planned dialogue paths as prompts to steer dialogue generation. Experiments show that our proposed method outperforms other methods significantly.

\section*{Limitations}
Though our proposed method exhibits superior performance, we also recognize its limitations and discuss potential solutions. Our proposed method for goal-directed dialogue generation suffers from error propagation since the three stages perform in a pipeline manner. After analyzing those generated utterances with low human evaluation scores, we find that the performance of dialogue generation is prone to drop when our \textsc{Color} model fails to plan an appropriate dialogue path. We intend to alleviate this issue by introducing some techniques in the cascaded generation, such as noisy channel models \cite{shannon1948mathematical,liu-etal-2021-pretraining}. In addition, other issues, such as how to make existing goal-directed dialogue systems more engaging and personalized, are worth further exploring.

\section*{Ethical Considerations}
Goal-directed dialogue systems can be used for creating non-obtrusive recommendations for specific products and services, introducing interesting new topics and educating users about those topics, and so forth. Developing such systems requires careful consideration since it has a broad impact on applications. The intention of our work is not to force the system to reach the designated target nor force users to accept recommendations. Instead, we aim to build better assistive technologies to improve the proactiveness of dialogue systems. Furthermore, our experimental datasets are publicly available. They have been filtered for sensitive and private information during dataset construction.

We hope to raise awareness of the potential for misuse of such systems with toxic intentions. For example, such systems may be used to pose as humans and actively manipulate users' perceptions on specific issues or political inclinations. To mitigate these risks, we emphasize the importance of improving transparency through regulations. It is essential to inform users that they are conversing with a bot instead of a human, and regulations on target designation are crucial when deploying these systems in specific domains. It is necessary to ensure that setting a target does not violate factual accuracy, user privacy rules, or human laws.

\section*{Acknowledgments}
This work was supported by the Research Grants Council of Hong Kong (15207122, 15207920, 15207821, 15204018) and National Natural Science Foundation of China (62076212). It was also supported in part by PolyU internal grants (ZVQ0, ZVVX).

\bibliography{custom}
\bibliographystyle{acl_natbib}

\appendix

\section{Dataset Descriptions and Pre-processing}
\label{appendix:dataset}
\paragraph{DuRecDial 2.0}

The DuRecDial 2.0 \cite{liu-etal-2021-durecdial} dataset is collected from crowdsourced human-to-human dialogues. In each dialogue, one person is defined as the seeker (the user's role) and the other as the recommender (the system's role). The recommender needs to proactively lead the dialogue and make recommendations by introducing new topics. Each seeker is equipped with a user profile containing user attributes (e.g., age range) and his/her past preference information. In order to smoothly converse with the seeker, the recommender has a domain knowledge graph consisting of domain-specific topics (e.g., movies, music) with related attributes. More importantly, a dialogue path composed of dialogue actions and topics is annotated for the recommender from the beginning to the end of the dialogue. All dialogues are aligned across the English and Chinese languages. We adopt the dataset in English for experiments.

\begin{figure}[t!]
\centering
\includegraphics[width=0.92\linewidth]{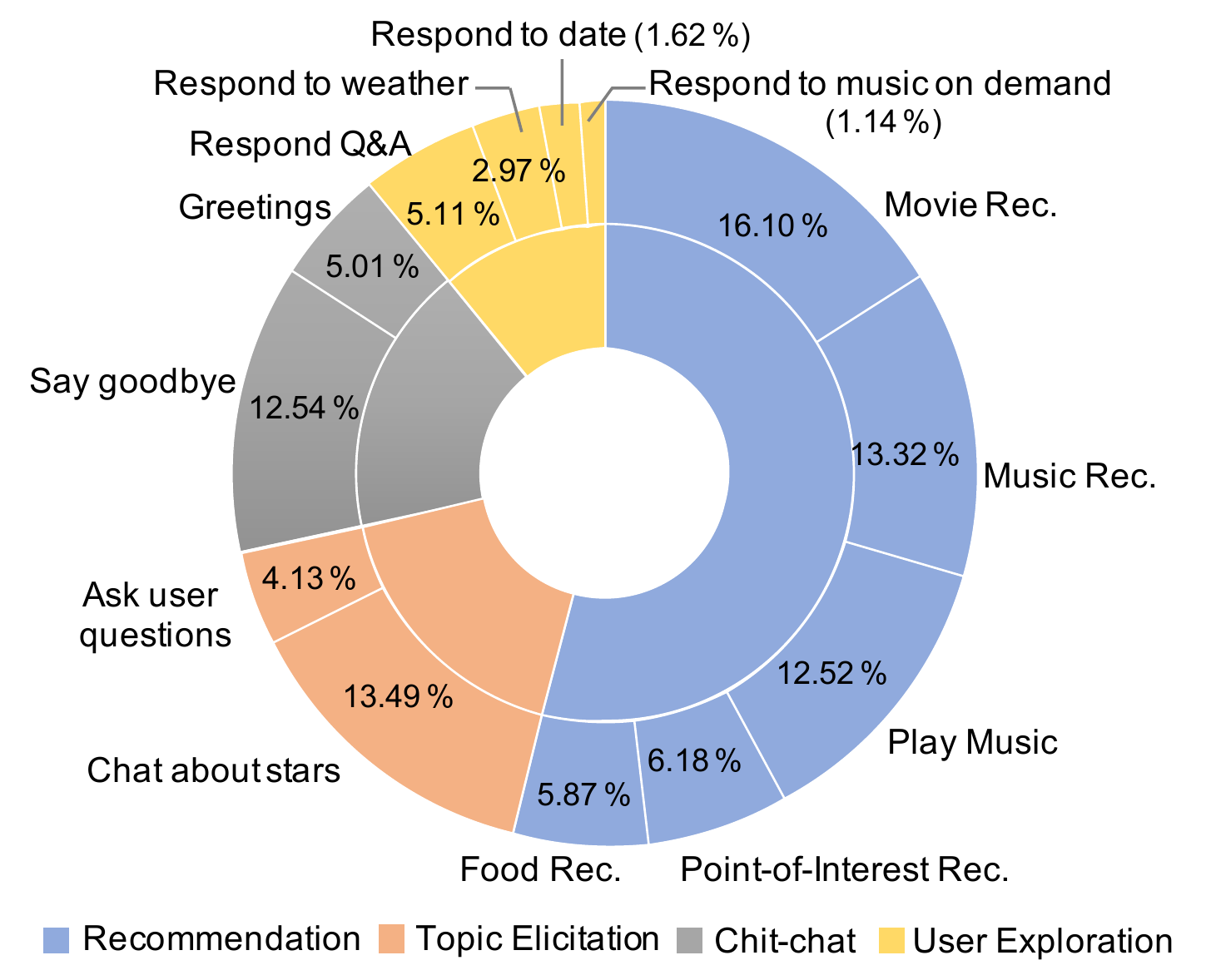}
\caption{Statistics of the system's dialogue actions on the repurposed DuRecDial 2.0 dataset.}
\label{fig:action_stats}
\end{figure}

Since there are no explicit annotated targets, we repurpose the original dataset automatically. For all those dialogues that are proactively led by the system, we treat the topic that the user has accepted at the end of each dialogue as the target topic, and view the system's corresponding action (e.g., movie recommendation, point-of-interest recommendation, etc.) as the target action. Each target topic is guaranteed to be grounded in the domain knowledge triples corresponding to the dialogue. We filter out those dialogues without introducing any new recommendation topics. The total number of topics is 628 (including a \texttt{NULL} topic). Figure \ref{fig:action_stats} shows the statistics of all the system's actions.  We observe an average of 4.3 $\sim$ 4.8 action-topic transitions from the beginning to reaching the target.

Following the splitting criterion \cite{liu-etal-2021-durecdial}, we obtain training/validation/test sets with 4,256/608/1,216 dialogues, respectively. To investigate the performance of different methods for goal-directed dialogue generation, we further use the dataset with two types of splits for the test set: \textit{in-domain (ID)} split and \textit{out-of-domain (OOD)} split, similar to \citet{sevegnani-etal-2021-otters,gupta-etal-2022-target}. The OOD split ensures that none of the target topics in the test set are present in the training set. In contrast, the target topics in the ID split are allowed to appear in the training set.

\paragraph{TGConv}
The TGConv \cite{yang-etal-2022-topkg} dataset is extracted based on the chit-chat corpus ConvAI2 \cite{dinan2020second}, and the external commonsense KG ConceptNet \cite{speer2017conceptnet}. In the TGConv dataset, all target-oriented samples are identified by the dialogue utterances containing a go-through keyword/concept sequence that aligns with the KG path over the ConceptNet. Suppose the designated global target keyword is $w_n$, a transition path of keywords or concepts $\mathcal{P}=\{w_1\rightarrow \cdots\rightarrow w_n\}$ is annotated for each dialogue. Here, each neighbor word pair (i.e., $w_i$ and $w_{i+1}$) is direct or low-order connected in the ConceptNet. On average, the number of transitions from the start context to the target is approximately 5.
Furthermore, the target keywords are distinguished into ``\textit{easy-to-reach}'' and ``\textit{hard-to-reach}''. Specifically, the \textit{easy-to-reach} targets refer to target keywords with high frequency in the corpus. In comparison, target words with low frequency (less than 800) in the corpus are classified as \textit{hard-to-reach} targets because there are fewer cases to learn the transition to low-frequency target words. In this work, we follow the same data splitting as in \citet{yang-etal-2022-topkg} for experiments.

\section{Baseline Methods}

\subsection{Dialogue Generation}
\label{appendix:baseline_gen}
To evaluate dialogue generation quality, we first consider the following PLMs-based methods:
\begin{itemize}[leftmargin=*]
\item \textbf{GPT-2} \cite{radford2019language}: It is an autoregressive generation model for language generation. We use the GPT-2 base\footnote{\url{https://huggingface.co/gpt2}} model for fine-tuning.
\item \textbf{DialoGPT} \cite{zhang-etal-2020-dialogpt}: It is an autoregressive dialogue generation model pre-trained using large-scale dialogue corpora. We adopt the pre-trained small\footnote{\url{https://huggingface.co/microsoft/DialoGPT-small}} model for fine-tuning.
\item \textbf{BART} \cite{lewis-etal-2020-bart}: It is a denoising encoder-decoder model for language generation. We use the BART-base\footnote{\url{https://huggingface.co/facebook/bart-base}} model for fine-tuning.
\end{itemize}
Note that these models concatenate all parts of input texts described in the problem formulation as the model input and are fine-tuned to generate utterances directly.

On the DuRecDial 2.0 dataset, we additionally consider several competitive models that follow the planning-enhanced generation paradigm:
\begin{itemize}[leftmargin=*]
\item \textbf{MGCG\_G} \cite{liu-etal-2020-towards-conversational}: It employs the predicted next dialogue action and next topic to guide utterance generation. We re-run the officially released code\footnote{\url{https://github.com/PaddlePaddle/Research/tree/master/NLP/ACL2020-DuRecDial}} on the repurposed dataset.
\item \textbf{KERS} \cite{zhang-etal-2021-kers-knowledge}: It leverages a knowledge-enhanced mechanism to guide dialogue generation. We re-run the  available code\footnote{\url{https://github.com/z562/KERS}} on the repurposed dataset.
\item \textbf{TCP-Dial} \cite{wang2022follow}: It builds a target-driven conversation planning method to explicitly extract necessary knowledge and then guides dialogue generation. We re-run the available code\footnote{\url{https://github.com/iwangjian/Plan4RecDial}} on the repurposed dataset.
\end{itemize}

On the TGConv dataset, we consider the following competitive models:
\begin{itemize}[leftmargin=*]
\item \textbf{MultiGen} \cite{ji-etal-2020-language}: It is a language generation model with multi-hop reasoning on commonsense knowledge graphs.
\item \textbf{DKRN} \cite{qin2020dynamic}: It builds a dynamic knowledge routing network for topic transitions.
\item \textbf{CKC} \cite{zhong2021keyword}: It is a keyword-guided neural conversational model that leverages ConceptNet for keyword transitions.
\item \textbf{TopKG} \cite{yang-etal-2022-topkg}: It employs global planning on ConcepNet to guide dialogue generation and is the state-of-the-art approach\footnote{\url{https://github.com/yyyyyyzt/topkgchat}} on the TGConv dataset.
\end{itemize}

\subsection{Dialogue Planning}
\label{appendix:baseline_plan}
To compare the performance of dialogue planning for goal-directed dialogues, we consider the following dialogue planning methods: 
\begin{itemize}[leftmargin=*]
\item \textbf{MGCG} \cite{liu-etal-2020-towards-conversational}: It makes multi-task predictions for the next-turn's dialogue action and topic. However, it assumes that ground-truth historical dialogue actions and topics are known for a system. For a fair comparison in this work, we adopt the same input as our problem definition to conduct multi-task predictions.
\item \textbf{KERS} \cite{zhang-etal-2021-kers-knowledge}: It generates the next-turn's dialogue action and topic based on a Transformer \cite{vaswani2017attention} network. 
\item \textbf{TCP} \cite{wang2022follow}: It is a target-driven planning framework that plans a dialogue path consisting of dialogue actions and topics in a generation-based manner.
\item \textbf{TopKG-Plan} \cite{yang-etal-2022-topkg}: It employs reinforcement learning to plan a commonsense keyword path based on ConceptNet.
\item \textbf{BERT} \cite{devlin-etal-2019-bert}: Based on the intuition of multi-task predictions, we fine-tune the widely-used BERT model by adding two fully-connected layers to jointly predict the next-turn's dialogue action and topic. We use the uncased BERT-base\footnote{\url{https://huggingface.co/bert-base-uncased}} model for fine-tuning.
\item \textbf{BART} \cite{lewis-etal-2020-bart}: Based on the intuition of generation, we use the BART-base model for fine-tuning, which is then used to generate a dialogue path similar to ours.
\end{itemize}

\section{Training and Inference Details}
\label{appendix:training}
In Stage 1, we set the batch size for contrastive training to 64 and adopt the Adam \cite{kingma2014adam} optimizer with a learning rate of $2e\text{-}4$. We train our encoder $\mathcal{F}$ for 10 epochs. For training in Stage 2, we adopt the Adam optimizer with an initial learning rate of $2e\text{-}5$ and warm up over the first 10\% training steps. We train our \textsc{Color} for a maximum of 10 epochs with a batch size of 16. The best checkpoint is chosen based on its performance on the validation set. For inference, we employ greedy decoding to generate a dialogue path token by token, with a maximum decoding length of 80. In Stage 3, we employ GPT-2 base and DialoGPT-small (see Appendix \ref{appendix:baseline_gen}) as our backbone models. We follow the description in \S \ref{sec:sec3} and fine-tune backbone models for 10 epochs. For a fair comparison, we use greedy decoding with a maximum decoding length of 100 for all models. We conduct experiments on one NVIDIA 3090 GPU machine.

\section{Procedure of Human Evaluation}
\label{appendix:humaneval}
For turn-level evaluation, we randomly sampled 50 dialogues from the ID test split and 50 dialogues from the OOD test split from the DuRecDial 2.0 dataset. We then compared the generated utterances of the following models: MGCG\_G, DialoGPT, TCP-Dial, and ours (\textsc{Color} w/ DialoGPT). For a fair comparison, the models were randomly renamed as ``model-1'', ``model-2'', and so forth. The annotators were then asked to mark scores for the compared models from (1) appropriateness (\textbf{Appr.}), which measures whether the utterance responds to the dialogue context appropriately, and (2) informativeness (\textbf{Info.}), which measures whether the utterance is informative by making full use of the grounded knowledge. 

For dialogue-level evaluation, we asked our annotators to act as users and converse with the models. Each model's generated utterance in the current turn was used as part of the dialogue context in the next turn. Our annotators were asked to maintain consistency with the equipped user profile. To ensure diverse evaluation targets, we randomly selected 5 target actions from the DuRecDial 2.0 test set, each paired with 10 different target topics, resulting in a total of 50 targets evaluated. We did not expose the targets to them during human-model conversations and restricted all conversations to no more than 12 turns. We finally released the designated targets to the annotators and asked them to score the models based on three evaluation metrics: (1) proactivity (\textbf{Proact.}), which measures whether a model proactively leads the dialogue; (2) coherence (\textbf{Coh.}), which manually examines whether the whole dialogue is fluent, coherent, and smoothly transited; (3) goal success (\textbf{Succ.}), which estimates whether a model effectively reaches the target.

Our annotators were required to score the generated dialogues rating in $\{0, 1, 2\}$, where higher is better. The agreement among the annotators is measured by Fleiss's kappa \cite{fleiss1971measuring}. We reported each model's average score from different annotators as the ultimate human evaluation result. In addition, we transparently informed all annotators of our research intent. We paid reasonable wages and provided enough time for the annotators to complete the evaluation.

\section{Case Study}
\label{appendix:case_study}

Table \ref{tab:case_durecdial} and Table \ref{tab:case_tgconv} show some cases on the DuRecDial 2.0 and TGConv datasets, respectively.

\begin{table*}[!ht]
\small
\addtolength{\tabcolsep}{-3pt}
\begin{center}
\begin{tabular}{L{1.4cm} L{7.8cm} L{6.1cm}}
\toprule
\multicolumn{3}{c}{\textbf{Target}: \bluebox{Movie recommendation} (target action) | \greenbox{The Art of Action: Martial Arts in Motion} (target topic)} \\
\toprule
\textbf{Model} & \textbf{Example} & \textbf{Dialogue Planning} \\

\midrule[.04em]
MGCG\_G
& 
\textbf{User:} Do you know who starred in the movie \underline{Jackie Chan:} \underline{My Story}? \par
\textbf{Bot:} of course. it’s jackie chan. \par
\textbf{User:} Well, I see. You are so brilliant! \par
\textbf{Bot:} you’re welcome. \redbox{let’s talk about movies. let’s talk about} \redbox{movies.} (\textcolor{red}{not fluent and coherent})\par
\textbf{User:} OK. What would you like to talk about? \par
\textbf{Bot:} \underline{jackie chan}'s movies are very good. \par
\textbf{User:} Yeah, that’s why I love him so much. \par
\textbf{Bot:} \redbox{i recommend you to see \underline{rumble in the bronx}}. it’s very good. it’s a very good movie. \par
\textbf{User:} Yeah. Of course. \par
\textbf{Bot:} you have a good taste. \redbox{the \underline{rumble in the bronx} is} \redbox{recommended to you}. \par
... (\textcolor{red}{not reach the target})

& 
Respond Q\&A | Jackie Chan: My Story \par
~ \par
Respond Q\&A | Jackie Chan: My Story \par
~ \par
~ \par
Chat about stars | Jackie Chan \par
~ \par
\bluebox{Movie recommendation} | \redbox{Rumble in the Bronx} \par
~ \par
~ \par
\bluebox{Movie recommendation} | \redbox{Rumble in the Bronx} \par
~ 
\\

\midrule[.02em]

DialoGPT
&
\textbf{User:}  Do you know who starred in the movie \underline{Jackie Chan:} \underline{My Story}? \par
\textbf{Bot:} Yes, it's \underline{Jackie Chan}. \par
\textbf{User:} Well, I see. You are so brilliant! \par
\textbf{Bot:} You're welcome. Speaking of \underline{Jackie Chan}, he is really a very good actor. \par
\textbf{User:} Yeah. I sincerely love him! \par
\textbf{Bot:} Yes, he has won the \underline{Golden Horse Award for Best} \underline{Actor}. \par
\textbf{User:} Wow, that’s great. \par
\textbf{Bot:} He also won the \underline{Hong Kong Film Award for Best Actor}. \par
\textbf{User:} Anything else? \par
\textbf{Bot:} Well, \redbox{it's a \underline{documentary} movie. It's worth to} \redbox{recommend to you}. Don't miss it! (\textcolor{red}{not coherent with context}) \par
... (\textcolor{red}{get trapped})

& 
~~~~~~(no dialogue planning)
\\

\midrule[.02em]

Ours
&
\textbf{User:} Do you know who starred in the movie \underline{Jackie Chan:} \underline{My Story}? \par
\textbf{Bot:} Yes, it’s \underline{Jackie Chan}. \par
\textbf{User:} Well, I see. You are so brilliant! \par
\textbf{Bot:} You’re welcome. \underline{Jackie Chan} is a Chinese martial arts superstar. \par
\textbf{User:} Yeah. I sincerely love him! \par
\textbf{Bot:} Well, he has \greenbox{won the \underline{Golden Horse Award for Best}} \greenbox{\underline{Actor}. He has many good movies}. (\textcolor{blue}{more coherent}) \par
\textbf{User:} Wow, that’s great. \par
\textbf{Bot:} I'd recommend his movie \underline{The Art of Action: Martial} \underline{Arts in Motion Picture} to you. \par
\textbf{User:} Sounds good. \par
\textbf{Bot:} It's a \underline{documentary} movie expressing Hong Kong \underline{kung fu} movies. Don't miss it. 
\par 
(\textcolor{blue}{succeed to reach the target})

& 
$\mathcal{P}_1$: \texttt{[A]} Respond Q\&A \texttt{[T]} Jackie Chan: My Story \texttt{[A]} Chat about stars \texttt{[T]} Jackie Chan \texttt{[A]} Movie recommendation \texttt{[T]} The Art of Action: Martial Arts in Motion Picture \par

$\mathcal{P}_2$: \texttt{[A]} Chat about stars \texttt{[T]} Jackie Chan \texttt{[A]} Movie recommendation \texttt{[T]} The Art of Action: Martial Arts in Motion Picture \par
~ \par
$\mathcal{P}_3$: (same as $\mathcal{P}_2$) \par
~ \par

$\mathcal{P}_4$: \texttt{[A]} \bluebox{Movie recommendation} \texttt{[T]} \greenbox{The Art} \greenbox{of Action: Martial Arts in Motion Picture} \par
~ \par
$\mathcal{P}_5$: (same as $\mathcal{P}_4$) \par
~ \par
~\par
\\
\toprule
\end{tabular}
\end{center}
\caption{Generated cases from dialogue-level human evaluations on the DuRecDial 2.0 dataset. The bot's utterances are generated by different models, including (a) MGCG\_G, (b) DialoGPT, and (c) Ours (\textsc{Color} w/ DialoGPT). The topics and topic-related attributes that also appear in the domain knowledge are marked with underlines.}
\label{tab:case_durecdial}
\end{table*}

\begin{table*}[!ht]
\small
\addtolength{\tabcolsep}{-3pt}
\begin{center}
\begin{tabular}{L{1.2cm} L{7.6cm} L{6.6cm}}
\toprule
\multicolumn{3}{l}{
\textbf{Start Keywords:} wear $\rightarrow$ shoes ~~~~~~~~
\textbf{Easy Target}: \greenbox{computer}} \\
\textbf{Context}  &  
\textbf{A:} it is freezing outside, i enjoy \redbox{wearing} skinny jeans. \par
\textbf{B:} i am often gunning for new \redbox{shoes} lol. 
& \\

\midrule[.02em]

\textbf{Model} & \textbf{Example} & \textbf{Dialogue Planning} \\

\midrule[.02em]

TopKG
&
\textbf{A:} i love tee \redbox{shirts} with rock band logos.
\par
\textbf{B:} that sounds more cozy. what is your favorite \redbox{band}?
\par
\textbf{A:} i'm young but i love old \redbox{stuff}, beatles.
\par
\textbf{B:} that is nice.
\par
\textbf{A:} it's cool! i like working on the \greenbox{computer}
\par
(\textcolor{blue}{succeed to reach the target})

&  shirt $\rightarrow$ band $\rightarrow$ stuff $\rightarrow$ computer

\\

\midrule[.02em]

Ours
&
\textbf{A:} oh, me too. my favorite \redbox{shirts} are flannel ruffle sleeves.
\par
\textbf{B:} really that's sick love them, why do you \redbox{like} that?
\par
\textbf{A:} i enjoy \redbox{playing} video \redbox{games} and listening to music.
\par
\textbf{B:} is that \redbox{game} rolls royce's ghost tho am i right.
\par
\textbf{A:} yeah, it is the wallpaper for my \greenbox{computer}.
\par
(\textcolor{blue}{succeed to reach the target more coherently})
\par
~
& 
$\mathcal{P}_1$: \texttt{[T]} shirt \texttt{[T]} band \texttt{[T]} music \texttt{[T]} computer
\par
$\mathcal{P}_2$: \texttt{[T]} like \texttt{[T]} music \texttt{[T]} computer
\par
$\mathcal{P}_3$: \texttt{[T]} play \texttt{[T]} game \texttt{[T]} computer
\par
$\mathcal{P}_4$: \texttt{[T]} game \texttt{[T]} computer
\par
$\mathcal{P}_5$: \texttt{[T]} computer
\par
~

\\
\midrule
\midrule

\multicolumn{3}{l}{
\textbf{Start Keywords:}  day $\rightarrow$ living ~~~~~~~~
\textbf{Hard Target}: \greenbox{barbershop}} \\
\textbf{Context}  &  
\textbf{A:} i enjoy lawn mowing on sunny \redbox{days}.
\par
\textbf{B:} i enjoy green \redbox{living} and just put solar panels on roof.
& \\
\midrule[.02em]
\textbf{Model} & \textbf{Example} & \textbf{Dialogue Planning} \\
\midrule[.02em]

TopKG
&
\textbf{A:} i do live rpg \redbox{events} over the weekends.
\par
\textbf{B:} wow, are you living in \redbox{game} or like that?
\par
\textbf{A:} i gonna make it a big one \redbox{day}.
\par
\textbf{B:} amazing \redbox{dream}, huh?
\par
... (\textcolor{red}{get trapped})

&  
event $\rightarrow$ game $\rightarrow$ day $\rightarrow$ dream $\rightarrow$ $\cdots$

\\
\midrule[.02em]

Ours
&
\textbf{A:} cool. do you only \redbox{live} with your parents?
\par
\textbf{B:} yes i do. i also have two \redbox{dogs} myself.
\par
\textbf{A:} ha, i love dogs. but my italian \redbox{wife} doesn't.
\par
\textbf{B:} not easy. dogs must spend a lot of time \redbox{outdoors}.
\par
\textbf{A:} lol, but she love going to the pet \greenbox{barbershop} for pictures.
\par
(\textcolor{blue}{succeed to reach the target})

& 
$\mathcal{P}_1$: \texttt{[T]} live \texttt{[T]} family \texttt{[T]} outdoor \texttt{[T]} barbershop
\par
$\mathcal{P}_2$: \texttt{[T]} dog \texttt{[T]} room \texttt{[T]} outdoor \texttt{[T]} barbershop
\par
$\mathcal{P}_3$: \texttt{[T]} wife \texttt{[T]} outdoor \texttt{[T]} barbershop
\par
$\mathcal{P}_4$: \texttt{[T]} outdoor \texttt{[T]} barbershop
\par
$\mathcal{P}_5$: \texttt{[T]} outdoor \texttt{[T]} barbershop
\par
~

\\
\bottomrule
\end{tabular}
\end{center}
\caption{Generated cases from self-play simulations on the TGConv dataset. The compared models include (a) TopKG and (b) Ours (\textsc{Color} w/ DialoGPT).}
\label{tab:case_tgconv}
\end{table*}

\end{document}